\documentclass[final,3p,times]{elsarticle}

\usepackage{setspace}
\usepackage{times}
\usepackage{epsfig}
\usepackage{graphicx}
\usepackage{amsmath}
\usepackage{amssymb}

\usepackage{bigstrut}
\usepackage{subcaption}
\usepackage{bm}
\usepackage{algorithm,lipsum}
\usepackage{wrapfig}
\usepackage[noend]{algpseudocode}

\usepackage[pagebackref=true,breaklinks=true,colorlinks,bookmarks=false]{hyperref}

\newcommand{\myparagraph}[1]{\vspace{6pt}\noindent{\bf #1}}

\begin{document}
	
	\begin{frontmatter}
		
		
		\author[label1]{Chandan Gautam}
		\ead{chandang@iisc.ac.in chandangautam31@gmail.com}
		\author[label1]{Sethupathy Parameswaran}
		\ead{sethupathyp@iisc.ac.in}
		\author[label2]{Ashish Mishra}
		\ead{mishra@cse.iitm.ac.in}
		\author[label1]{Suresh Sundaram\corref{cor1}}
		\ead{vssuresh@iisc.ac.in}
		
		\cortext[cor1]{Corresponding author}
		\address[label1]{Indian Institute of Science, Bangalore}
		\address[label2]{Indian Institute of Technology Madras}
		
		\title{}
		\title{Generative Replay-based Continual Zero-Shot Learning}

		\address{}

		\begin{abstract}
			Zero-shot learning is a new paradigm to classify objects from classes that are not available at training time. Zero-shot learning (ZSL) methods have attracted considerable attention in recent years because of their ability to classify unseen/novel class examples. Most of the existing approaches on ZSL  works when all the samples from seen classes are available to train the model, which does not suit real life. In this paper, we tackle this hindrance by developing a generative replay-based continual ZSL (GRCZSL). The proposed method endows traditional ZSL to learn from streaming data and acquire new knowledge without forgetting the previous tasks' gained experience. We handle catastrophic forgetting in GRCZSL by replaying the synthetic samples of seen classes, which have appeared in the earlier tasks. These synthetic samples are synthesized using the trained conditional variational autoencoder (VAE) over the immediate past task.
			Moreover, we only require the current and immediate previous VAE at any time for training and testing.  The proposed GRZSL method is developed for a single-head setting of continual learning,  simulating a real-world problem setting. In this setting, task identity is given during training but unavailable during testing. GRCZSL performance is evaluated on five benchmark datasets for the generalized setup of ZSL with fixed and dynamic (incremental class) settings of continual learning. The existing class setting presented recently in the literature is not suitable for a class-incremental setting. Therefore, this paper proposes a new setting to address this issue. Experimental results show that the proposed method significantly outperforms the baseline and the state-of-the-art method and makes it more suitable for real-world applications.
		\end{abstract}
		
	\end{frontmatter}
	\section{Introduction}
	Image classification using deep learning has exhibited satisfactory performance for the fully supervised learning task. However, these conventional deep learning algorithms rely heavily on a large amount of labeled visual samples. However, it is challenging to collect labeled samples for every class in the real world. It leads to a problem of recognizing samples from unseen/novel classes while there are no visual samples present in the training data. It is an easier task to recognize unseen classes based on the unseen class descriptions by a human. Humans can do this based on their imagination capability, which they have learned from their past experiences. The solution approach in machine learning literature is popularly known as zero-shot learning (ZSL)~\cite{aPY,AWA,zeroshotlearning_dataset}. ZSL receives a surge of interest in the research community in the past years, and the researchers have developed various methods to tackle this problem. Despite the development of various ZSL methods, it is difficult to continuously learn from the sequence of tasks without forgetting the previously accumulated knowledge.  Leveraging past experience and acquiring new knowledge from the streaming data is known as continual/lifelong learning. This learning strategy needs to merge with ZSL to enable ZSL for a continual learning framework, i.e., continual zero-shot learning (CZSL).
	
	In conventional ZSL, the purpose of the trained model is to identify the unseen classes in two kinds of settings during testing; (i) disjoint setting: classification search space consists of only unseen classes, (ii) generalized setting: classification search space consists of seen and unseen classes both. Here, seen and unseen classes are completely disjoint, i.e., seen classes $\cap$ unseen classes $= \phi$. A generalized setting is more realistic than a disjoint setting as a model doesn't know during testing whether an upcoming image belongs to a seen or unseen class. This paper also follows the generalized setting as it is closer to the real-world problem. Generally, ZSL can be performed in two ways; (i) by learning to map from the visual space to the semantic space and vice-versa using a mapping function, (ii) by generating synthetic samples for unseen classes using a generative model. However, these ZSL methods are not effective in handling the CZSL problem.
	
	Most recently, only a handful of researchers have made an effort to tackle this CZSL problem \cite{wei2020lifelong,skorokhodov2020normalization}. CZSL has been developed for two kinds of continual learning setting, i.e., multi-head \cite{wei2020lifelong} and single-head \cite{skorokhodov2020normalization} setting. In a multi-head setting, task identity is known during training and testing, and there is a separate classifier for each task. In a single-head setting, task identity is not known during inference time, and a shared classifier is used among all tasks. It can be stated that the assumption of a multi-head setting is not practical and not a realistic setting. However, the single-head setting is closer to the real-world setting. Therefore, we also develop a single-head setting-based continual learning method for generalized ZSL. The existing single-head setting \cite{skorokhodov2020normalization} is not suitable for class-incremental learning as it assumes that semantic information of all seen and unseen classes are known at the very first task. It is an unrealistic assumption. Therefore, we propose a new setting for addressing this issue. This proposed CZSL setting also helps analyze the model's performance compared to the offline mode (i.e., upper bound). In this paper, for developing a CZSL method, we need to deal with two aspects: continual learning and ZSL. Continual learning is dealt with using a generative replay, and ZSL is dealt with using a generative model conditional variational autoencoder (CVAE) \cite{mishra2018generative}. 
	
	The overall contribution of this work can be summarized as follows:
	\begin{enumerate}
		\item The existing generative replay-based method is developed for performing a traditional classification task (i.e., classification of seen classes) in a continual learning manner. However, this paper develops a generative replay-based continual ZSL (GRCZSL), which can also identify the unseen class by only using the class's semantic information.
		
		\item GRCZSL uses the immediate previous decoder model to handle catastrophic forgetting in CZSL, which makes it quite efficient. It does not require storing anything other than the previous decoder model.
		
		\item  GRCZSL has experimented on two kinds of continual learning settings. One CZSL setting already exists \cite{skorokhodov2020normalization}. The other is proposed for two purposes: (a) to evaluate the class-incremental ability of the CZSL method as the incremental class evaluation is not possible on the existing setting \cite{skorokhodov2020normalization}; (b) to compare the performance of the continual model from the offline model as the train-test split of the last task of the proposed CZSL setting is identical to standard ZSL split.
		
		\item As GRCZSL doesn't require a task-information at the inference time, it performs task-agnostic prediction and is suitable for the single-head setting. It shares the same model among all tasks.
		
		\item We conduct experiments on the five standard ZSL datasets by splitting these among different tasks for CZSL. Moreover, two kinds of CZSL settings are used for the experiment. GRCZSL outperforms existing state of the art method \cite{skorokhodov2020normalization} by more than $2\%$ and $4\%$. 		     
	\end{enumerate}
	
	The remaining paper is organized as follows:
	Section \ref{lit_survey} contains literature survey, Section \ref{pro_meth} provides the proposed method, experimental results and ablation study are discussed in Section \ref{perf_eval}, and the last section (i.e., Section \ref{concl}) concludes this paper.   
	\section{Related Work} \label{lit_survey}
	This section contains related work in three parts: (i) zero-shot learning (ii) continual learning, and (iii) continual zero-shot learning 
	
	\subsection{Zero-shot Learning}
	
	ZSL is initially introduced in \cite{AWA1} for attribute-based classification and considered a disjoint setting of ZSL. Generally, a traditional method learns an embedding by mapping from visual to semantic information during training. Then at testing time, predict a semantic vector for unseen class samples and assign a class based on the nearest neighbor algorithm \cite{socher2013zero,mishra2018generative,verma2017simple,wang2016relational}. Here, this kind of embedding can be learned by simply learning linear compatibility between the visual and semantic space \cite{akata2016label,DEVISE,SJE}. Further, a novel regularization-based method is proposed for learning this      
	linear compatibility function \cite{ALE,ConSE,lampert2014attribute}. Further, a bilinear compatibility function-based ZSL methods \cite{zhang2016zero,ESZSL2015,SAE} are also developed to improve the performance over linear one. In \cite{ESZSL2015}, they use a simple compatibility function to model the relationship between image and attribute feature space, as well as regularize the objective function explicitly. In \cite{SAE}, a method is developed by mapping visual to semantic and again mapping from mapped semantic space to visual space using an autoencoder loss to improve the reconstruction ability of the model. Generally, a linear model has a lot of bias and isn't often enough to model the image classification problem. It requires a non-linear decision boundary to correctly classify the images. As a result, non-linear compatibility learning techniques have been developed \cite{socher2013zero,xian2016latent}. In \cite{socher2013zero}, a cross-modal transfer-based method is developed for generalized ZSL, which uses a single-hidden layer-based neural network with a non-linear activation function. In \cite{xian2016latent}, a latent embedding-based method is proposed, which learns a non-linear compatibility function.

	In recent years, ZSL employs generative models such as variational autoencoder \cite{vae} and generative adversarial network \cite{gan,wgan} to learn the samples from attributes. These methods received significant attention in the literature due to its ability to generate high-quality synthetic samples. In literature various ZSL and GZSL method has been problem in recent years \cite{kumar2018generalized,verma2021towards,xian2019f,huang2019generative,sariyildiz2019gradient,mishra2018generative,eccv2018,schonfeld2019generalized,chou2021adaptive,keshari2020generalized}, and are significantly outperformed non-generative ZSL methods.   
	
	First, these generative methods generate synthetic samples for seen and unseen classes, and then a supervised classifier is trained based on these synthetic samples. In this approach, bias towards seen classes is handled effectively as we can generate as many samples as required for seen and unseen classes. We also use a generative method, i.e., VAE, as a base method for in this work.  
	
	\subsection{Continual Learning}
	
	There are mainly two issues in continual learning which need to be addressed during learning: catastrophic forgetting and intransigence. The whole work of continual learning can be categorized into two parts: (i) regularization-based methods \cite{kirkpatrick2017overcoming,rebuffi2017icarl,chaudhry2018riemannian} (ii) replay-based methods \cite{lopez2017gradient,shin2017continual,chaudhry2018efficient,hayes2019memory,chaudhry2019tiny}. Few regularization-based methods handle catastrophic forgetting by storing previously learned networks and performing knowledge distillation using it \cite{rebuffi2017icarl}. While other regularization-based methods control the parameters and discourage the modification of the learned parameters on the previous tasks to avoid catastrophic forgetting \cite{kirkpatrick2017overcoming}. Replay-based methods either replay the samples from the past task while training the current task or use those samples to provide the constraint on the optimization problem in such a way so that loss on the previous tasks must not increase \cite{lopez2017gradient,chaudhry2018efficient}. Since replay-based methods outperform regularization-based methods in the literature \cite{shin2017continual}, we have also employed a generative replay-based strategy for handling catastrophic forgetting for CZSL.         
	
	\subsection{Continual Zero-shot Learning}
	
	Zero-shot learning from streaming data is an unexplored area of research. Only a handful of research is available for continual ZSL (CZSL) \cite{wei2020lifelong,skorokhodov2020normalization}. First, Chaudhry et al. \cite{chaudhry2018efficient} discussed the possibility of continual learning using average gradient episodic memory (A-GEM) in the multi-head setting. Recently, Wei et al. \cite{wei2020lifelong} developed a CZSL method using a generative model and knowledge distillation. However, this CZSL method is only compatible with multi-head settings. Most recently, A-GEM is further developed for single-head setting \cite{skorokhodov2020normalization} using episodic gradient memory.
	
	\section{Proposed Method: Generative Replay-based Continual Zero-shot Learning} \label{pro_meth}
	
	This section presents the problem formulation for generalized CZSL, then describes the proposed CVAE architecture used with the proposed Generative Replay-based Continual Zero-Shot Learning (GRCZSL) method in the subsequent subsection. In the last subsection, incremental class learning of the GRCZSL method is described.   
	
	\subsection{Problem Formulation}
	For $n_{tr}$ number of training samples in the $t^{th}$ task, the CZSL is trained on a data stream  $\mathcal{D}^t_{ \tau r} = \{(x_i^t, \iota_i^t, y_i^t, a_i^t)_{i=1}^{n_{\tau r}}\}$. Here, $x_i^t$, $\iota_i^t$, $y_i^t$, and $a_i^t$ denote feature vector, task identity, class label, and class attribute information for $i^{th}$ sample of $t^{th}$ task, respectively. This class attribute information is the key information which is required for performing CZSL. Further, CZSL is tested on the data stream $\mathcal{D}^t_{\tau s} = \{(x_i^t, y_i^t)_{i=1}^{n_{\tau s}}\}$. Here, $x_i^t$ and $y_i^t$ denote feature vector and class label for $i^{th}$ sample of $t^{th}$ task, respectively. $n_{ts}$ denotes number of test samples in $t^{th}$ task. Here, the objective is to develop an algorithm, which can perform continual zero-shot learning based on these notations.

	\subsection{Architecture of proposed Conditional Varitional Auto-Encoder}\label{subsec:cvae}
	
	\begin{figure*}[!h]
		\begin{centering}
			\includegraphics[width=16cm, height=8cm]{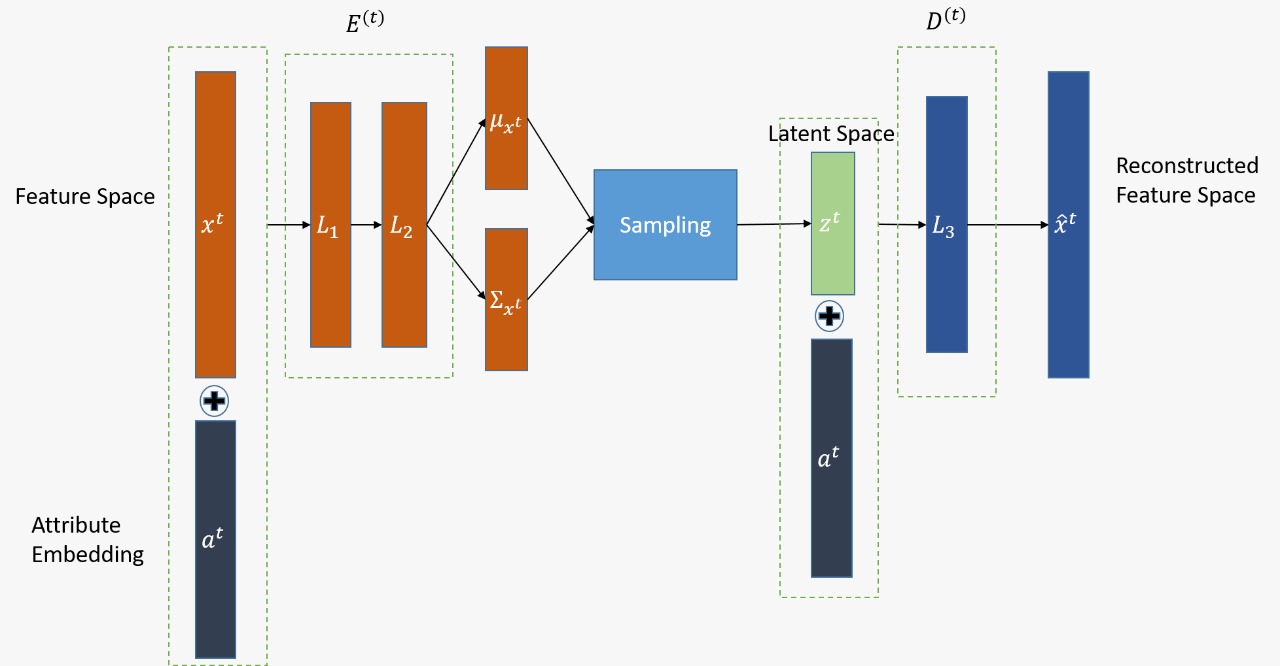}
			\par\end{centering}
		\caption{Network Architecture of CVAE used in the proposed generative replay-based CZSL (GRCZSL) method: The input feature $x$ and the respective attribute vector $a_y$ are concatenated and passed through a fully connected layer ($L_1$) of 512 units, followed by dropout with a dropout rate of 0.3 and passed through the fully connected layer ($L_2$) of 512 units. From ($L_2$), $\mu_z$ and $\Sigma_z$ are further obtained individually via another fully connected layer of dimension 50. A $z$ is sampled from the variational distribution $\mathcal{N}(\mu_x,\Sigma_x)$. The sampled $z$ concatenated with $a_y$ is passed to a fully connected layer ($L_3$) of 1024 units and then to the image space to reconstruct the original $x$. All activations are ReLU except the outputs of encoder and decoder which are linear.} 
		\label{net}
	\end{figure*}
	
	For developing the proposed method Generative Reply based Continual Zero-Shot Learning (GRCZSL), we have employed Variational Autoencoder (VAE) \cite{vae}. The VAE is a type of generative models which approximate the distribution of the latent space $z$ to that of the input data $x$, i.e., assume that there exists a hidden variable, say $z$, which can generate an observation $x$. Since, one can only observe $x$, we are interested in the prior $p(z|x)$ given by
	\begin{equation}
	p(z|x) = \frac{p(x|z) p(z)}{p(x)}
	\end{equation}
	
	But estimating $p(x)$ is quite difficult and it is often an intractable distribution. Hence using variational inference we approximate $p(z|x)$ using the parametrized distribution $q_{\Phi}(z|x)$. Since we want $q_{\Phi}(z|x)$ to be similar to $p(z)$, we can estimate the parameters of $q_{\Phi}(z|x)$ by minimizing the Kullback–Leibler (KL) divergence \cite{kullback1951information} loss ($\mathcal{L}_{KL}$) between the two distributions, which is computed as follows  
	
	\begin{equation}
	\mathcal{L}_{KL} = -KL\Big(q_{\Phi}(z|x)||p_{\theta}(z)\Big)
	\end{equation}
	
	Along with this KL divergence loss, VAE also minimizes the reconstruction loss ($\mathcal{L}_{Re}$):
	
	\begin{equation}
	\mathcal{L}_{Re} = \mathbb{E}_{q_{\Phi}(z)}\left[\log p_{\theta}(x|z)\right]
	\end{equation}
	
	Overall, VAE consists of the following loss, which is also called as variational lower bound for VAE:
	\begin{equation}\label{VAE}
	\mathcal{L}_{VAE} = -KL(q_{\Phi}(z|x)||p_{\theta}(z))
	+\mathbb{E}_{q_{\Phi}(z)}\left[\log p_{\theta}(x|z)\right]
	\end{equation}
	
	Here, $p_{\theta}(x|z)$ can be modelled as a decoder with parameters $\theta$ mapping latent space to data space and $q_{\Phi}(z|x)$ can be modelled as an encoder mapping data space to latent space. 
	
	The encoder gives the probability distribution $q(z|x,a)$, which is assumed to be an isotropic Gaussian distribution. We model the encoder using a neural network. The encoder takes the concatenated input feature $x$ and the respective attribute feature $a$ as input and gives the parameter vector of the Gaussian ($\mu_x, \Sigma_x$) as output. The decoder takes concatenated $z$ and $a$ and tries to reconstruct the $x$ of class $y$, which is most likely under the latent variable $z$. The decoder is modelled using a neural network. Once CVAE is properly trained, the decoder can be used to generate samples of any particular class, say $y$, by sampling $z$ from standard normal Gaussian and concatenating it with the respective $a$ and passing it to the decoder.

	Since VAE is a generative model, it can generate synthetic samples based on the learned distribution. However, it cannot perform conditional generation, i.e., generate samples for some specific class. For this purpose, a conditional VAE \cite{CVAE} has been developed and employed for ZSL \cite{mishra2018generative}.  CVAE maximizes the variational lower bound as follows:
	\begin{equation}
	\mathcal{L}_{CVAE} = -KL(q_{\Phi}(z|x,c)||p_{\theta}(z|c))
	+\mathbb{E}_{q_{\Phi}(z|c)}\left[\log p_{\theta}(x|z,c)\right]
	\end{equation}
	where $c$ is the condition. Here the class-specific attribute information $a$ is used as the condition. It allows us to generate samples based on the attributes $a$. It is to be noted that one wants $q(z)$ to be close to standard normal distribution. Hence, in this paper $p_{\theta}(z|c)$ is taken as $N(0,I)$.  
	
	Based on this CVAE, next, we present the proposed generative model-based CZSL method in the next section. 
	
	\subsection{Generative Replay-based CZSL: GRCZSL}
	\begin{figure*}[h!]
		\centering
		\includegraphics[width=1.0\linewidth]{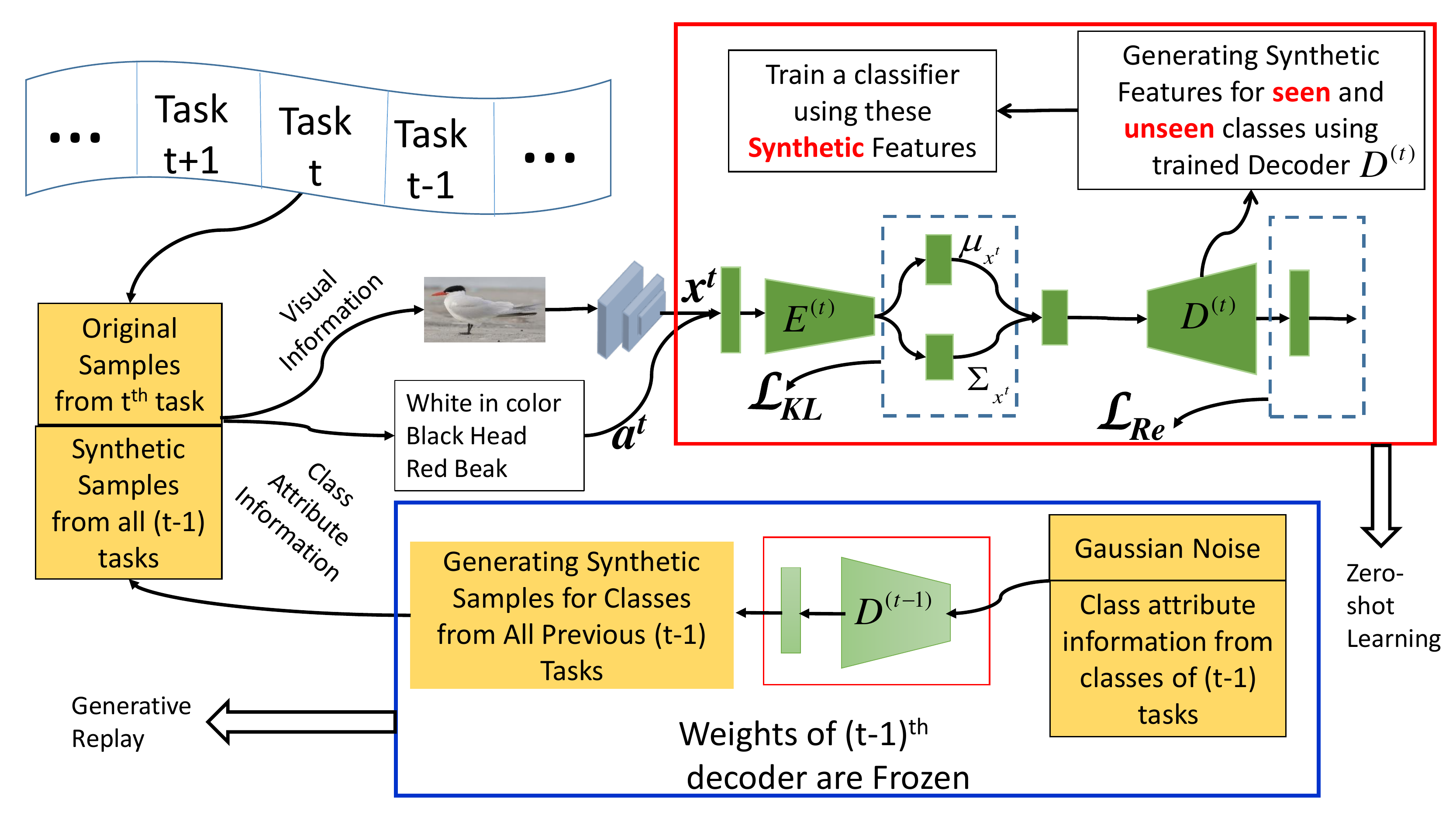}
		\caption{Flow diagram of GRCZSL. $E^{(t)}$ and $D^{(t)}$ are the encoder and decoder for $t^{th}$ task, which learns from the replay data and also performs CZSL. $D^{(t-1)}$ is the decoder of the $(t-1)^{th}$ task which generates sample for generative replay.  }
		\label{fig:GRCZSL}
	\end{figure*}

	In this paper, we propose a method for single-head setting, i.e., task information is not available at the inference time in the testing data stream (i.e., no task boundary). For this purpose, a generative replay-based CZSL (GRCZSL) method is proposed in this paper. A flow diagram of the GRCZSL method is depicted in Figure \ref{fig:GRCZSL}. The GRCZSL mainly has two components: (i) generative method for performing generalized ZSL, (ii) generative replay for performing continual learning with ZSL. A generative method is standard in both components, and CVAE is adapted for this purpose. However, one can use any generative model (architecture of CVAE is detailed in the above Section \ref{subsec:cvae}). GRCZSL first performs a traditional ZSL on the first task and employ CZSL from $2^{nd}$ task onward. 
	
	\myparagraph{For the first task:} GRCZSL is performed as follows: 
	\begin{enumerate}[(i)]
		\item Train the CVAE for the training samples of the seen classes at the first task. A CVAE \cite{CVAE} consists of an encoder ($E$) and decoder ($D$) network with parameters $\Phi$
		and $\theta$, respectively. It is a graphical model which estimates mean ($\mu_{x_{i}^1}$) and variance ($\Sigma_{x_{i}^1}$) from the input stream $x_i^1$ using an encoder, i.e., $q_{\Phi}(z_i^1|x_i^1, a_i^1)$. Here, $z_i^1$ represents latent variable, which is sampled from the estimated $\mathcal{N}(\mu_{x_{i}^1},\Sigma_{x_{i}^1})$. Further, latent variable $z_i^1$ is concatenated with class attribute $a_i^1$ and pass to the decoder $D$ and try to reconstruct the input $x_i^1$ at the output of the decoder. Overall, CVAE at the first task minimizes two losses simultaneously: Kullback–Leibler (KL) divergence \cite{kullback1951information} loss ($\mathcal{L}_{KL}$) at the latent space and reconstruction loss ($\mathcal{L}_{Re}$) at the output of decoder of CVAE as follows:
		\begin{equation}
		\begin{aligned}
		\mathcal{L}(\theta^1,\Phi^1;x^1, a^1) =\mathcal{L}_{Re}^1(x^1,\hat{x}^1)+		
		\mathcal{L}_{KL}^1\left(\mathcal{N}(\mu_{x^1},\Sigma_{x^1}),\mathcal{N}(0,I)\right),
		\end{aligned}
		\end{equation}
		where $\hat{x}^1$ is the predicted value by CVAE. For reconstruction loss, $L_2$ norm has been used.
		
		\item After training the CVAE model, generate synthetic samples for the seen and unseen classes using the decoder of the trained CVAE model. For this, Gaussian noise with the concatenation of seen and unseen classes attribute information is passed to the trained decoder. This trained decoder generates synthetic samples for each seen and unseen class as class attribute information.
		\item After generating synthetic samples for seen and unseen classes, a classifier is trained with these synthetic samples.
	\end{enumerate}             
	
	\myparagraph{For all tasks except first:} After training of the first task, the trained network from previous task $t-1$ is used to train for the new task $t$ as follows:
	\begin{enumerate}[(i)]
		\item For any $t^{th}$ task, weights of $(t-1)^{th}$ task's CVAE is frozen after the training.
		\item Now, the $t^{th}$ task's CVAE is initialized by the final weights of previous $(t-1)^{th}$ task's CVAE and it is trained.
		\item For training the $t^{th}$ task's CVAE, we pass the samples of seen classes at the $t^{th}$ task and the synthetic samples of the seen classes ($x'$) of all previous $(t-1)$ tasks, which are generated before training of $t^{th}$ CVAE. For generating these synthetic samples ($x'$) of classes of previous tasks, GRCZSL uses the trained decoder of $(t-1)^{th}$ task. These synthetic samples of the previous tasks helps GRCZSL in alleviating catastrophic forgetting of the CZSL.  
		\item Overall, $t^{th}$ task's CVAE minimizes the same two kinds of losses as first task (i.e., reconstruction loss ($\mathcal{L}_{Re}^t$) and KL divergence loss ($\mathcal{L}_{KL}^t$) as follows:
		\begin{equation}
		\begin{aligned}
		\mathcal{L}(\theta^t,\Phi^t;x^t, a^t) = & \alpha(\mathcal{L}_{Re}^t(x^t,\hat{x}^t)+ \mathcal{L}_{KL}^t\left(\mathcal{N}(\mu_{x^t},\Sigma_{x^t}),\mathcal{N}(0,I)\right)) +\\ &		 (1-\alpha)(\mathcal{L}_{Re}^t(x',\hat{x}')+
		+\mathcal{L}_{KL}^t\left(\mathcal{N}(\mu_{x'},\Sigma_{x'}),\mathcal{N}(0,I)\right)),
		\end{aligned}
		\end{equation}
		where $\alpha$ denotes task importance, $\mu_{x^t}$ and $\Sigma_{x^t}$ represent estimated mean and variance, respectively for the $t^{th}$ task using $t^{th}$ CVAE, and $\hat{x}'$ is the corresponding predicted value of $x'$ by the $t^{th}$ CVAE. Here, $t^{th}$ CVAE denotes trained CVAE model after $t^{th}$ task.   
		\item After training the CVAE for the $t^{th}$ task, we generate synthetic samples for seen and unseen classes by the decoder of the $t^{th}$ CVAE like the first task. At the end, the classifier is trained for the $t^{th}$ task using these synthetic samples (more detailed discussion is available in the next subsection, i.e., Section \ref{subsec:class_inc}). 
	\end{enumerate} 
	
	\subsubsection{Class Incremental Learning in GRCZSL}\label{subsec:class_inc}
	As discussed earlier, the CZSL problem is solved in a single-head setting, and it provides a task-agnostic prediction. If all training data are available apriori, then any classifier can learn all of these classes jointly. For the CZSL problem, training data arrives sequentially in the form of a task. Using the generative reply approach, GRCZSL generates synthetic features using $t^{th}$ decoder of GRCZSL for $t$ tasks and uses these synthetic features for classification. These synthetic features are beneficial as GRCZSL can generate as many samples as one wants for underrepresented classes. A single-hidden layer-based non-linear classifier using a softmax function is used as a classifier. Here, the classifier continually learns new classes as the next tasks are presented to the model.
	
	Overall, a CZSL method, namely, GRCZSL has been proposed. It requires only two networks, i.e., current (model at $t^{th}$ task) and previous (model at $(t-1)^{th}$ task) model, at any point of time during training a CZSL model. It is to be noted that only a decoder is required from the previous model as a decoder is sufficient to generate synthetic samples for previous tasks. Therefore, the proposed method is very efficient in terms of memory.
	
	\section{Performance Evaluation} \label{perf_eval}
	In this section, we evaluate the performance of GRCZSL using five ZSL benchmark datasets,  namely Attribute Pascal and Yahoo (aPY) \cite{aPY}, Animals with Attributes (AWA1 and AWA2) \cite{aPY}, Caltech-UCSD-Birds 200-2011 (CUB) \cite{CUB}, and SUN \cite{SUN}. The data are split into various tasks, and the details are given in Table \ref{tab_std_split}. For the purpose of performance comparison, we have used one of the standard settings and compared it with the existing state-of-the-art CZSL. Further, we have proposed another realistic setting to evaluate GRCZSL and compared it with the baseline. The respective performance metrics for different settings are described in the respective setting. All optimal hyperparameters are provided in Tables \ref{tab:GRCZSL_hyp_fixed} and \ref{tab:GRCZSL_hyp_dyn} for different settings. These settings and evaluation metrics are discussed in the subsequent subsection.
	
	\begin{table}[h]
		\centering
		\begin{tabular}{|l|p{1.6cm}|p{1cm}|p{1cm}|p{1cm}|}
			\hline
			Dataset & Attribute Dimension & Seen Classes & Unseen Classes & Total Classes   \\
			\hline
			CUB &  312 &  150 &  50 &  200\\
			aPY & 64 & 20 & 12 & 32\\
			AWA1  & 85 & 40 & 10 & 50\\
			AWA1  & 85 & 40 & 10 & 50\\
			SUN  & 102 & 645 & 72 & 717\\
			\hline
		\end{tabular}
		\caption{Standard Split of ZSL Datasets}
		\label{tab_std_split}
	\end{table}
	
	\begin{table*}[htbp]
		\centering
		\begin{tabular}{|l|c|c|c|c|c|}
			\hline
			Parameters & aPY & AWA1 & AWA2 & CUB & SUN \\
			\hline
			Learning rate(VAE) &  0.001 &  0.001 &  0.001 &  0.001 &  0.001\\
			Batch size (VAE) & 50 & 50 & 50 & 50 & 50\\
			Samples Generated per Seen Class (VAE) & 125 & 200 & 200 & 50 & 50\\
			Batch size of Generated Samples (VAE) & 15 & 20 & 50 & 100 & 100\\
			training epochs (VAE) & 25 & 25 & 25 & 25 & 25\\
			Hidden Neurons (classifier) & 1024 & 1024 & 1024 & 1024 & 512\\
			Learning rate (classifier) &  0.0001 &  0.0001 &  0.0001 &  0.0001 &  0.0001\\
			Weight Decay (classifier) & 0.001 & 0.001 & 0.001 & 0.001 & 0.001\\
			Batch size  (classifier) & 100 & 100 & 100 & 100 & 100\\
			training epochs (classifier) & 30 & 30 & 30 & 10 & 25\\	
			\hline	
		\end{tabular}
		\caption{Hyperparameters for GRCZSL in Fixed CZSL setting}
		\label{tab:GRCZSL_hyp_fixed}
	\end{table*}
	
	\begin{table*}[htbp]
		\centering
		\begin{tabular}{|l|c|c|c|c|c|}
			\hline
			Parameters & aPY & AWA1 & AWA2 & CUB & SUN \\
			\hline
			Learning rate(VAE) &  0.001 &  0.001 &  0.001 &  0.001 &  0.001\\
			Batch size (VAE) & 50 & 50 & 50 & 50 & 50\\
			Samples Generated per Seen Class (VAE) & 125 & 200 & 250 & 50 & 50\\
			Batch size of Generated Samples (VAE) & 15 & 800 & 20 & 100 & 100\\
			training epochs (VAE) & 25 & 25 & 25 & 25 & 25\\
			Hidden Neurons (classifier) & 1024 & 1024 & 1024 & 1024 & 512\\
			Learning rate (classifier) &  0.0001 &  0.0001 &  0.0001 &  0.0001 &  0.0001\\
			Weight Decay (classifier) & 0.001 & 0.001 & 0.001 & 0.001 & 0.001\\
			Batch size  (classifier) & 100 & 100 & 100 & 100 & 100\\
			training epochs (classifier) & 30 & 30 & 30 & 10 & 25\\	
			\hline	
		\end{tabular}
		\caption{Hyperparameters for GRCZSL in Dynamic CZSL setting.}
		\label{tab:GRCZSL_hyp_dyn}
	\end{table*}
	
	\subsection{Experimental Settings and Evaluation Metrics}
	
	Based on the assumption of seen and unseen classes for each task, two different experimental settings (fixed and dynamic settings) are designed. The details of each setting are given below:\\
	
	\myparagraph{Fixed CZSL Setting:} 
	
	In this setting, all classes till the current tasks are assumed as seen classes, and the classes from all remaining tasks are assumed as unseen classes. This setting is identical to the setting presented in \cite{skorokhodov2020normalization}. Following evaluation metrics are used for the performance analysis of this setting:  
	
	\begin{itemize}\itemsep-0.5em
		\item Mean Seen-class Accuracy (mSA)
		\begin{equation}
		mSA = \frac{1}{T}\sum_{t=1}^T CAcc(\mathcal{D}_{\tau s}^{\leq t}, A^{\leq t}),
		\end{equation}
		where $CAcc$ function computes per class accuracy, $T$ denotes the total number of tasks, $\mathcal{D}_{\tau s}$ denotes testing data, $\mathcal{D}^{\leq t}$ denotes all train/test data from $1^{st}$ to $t^{th}$ task, and $A^{\leq t}$ is the corresponding class attribute of those unseen classes.
		\item Mean Unseen-class Accuracy (mUA)
		\begin{equation}
		mUA = \frac{1}{T-1}\sum_{t=1}^{T-1} CAcc(\mathcal{D}_{\tau s}^{> t}, A^{> t}),
		\end{equation}
		where $\mathcal{D}^{> t}$ denotes all train/test data from $(t+1)^{th}$ to the last task, and $A^{> t}$ is the corresponding class attribute of those unseen classes.
		\item Mean Harmonic Accuracy (mH)
		\begin{equation}
		mH = \frac{1}{T-1}\sum_{t=1}^{T-1} H(\mathcal{D}_{\tau s}^{\leq t}, \mathcal{D}_{\tau s}^{> t}, A),
		\end{equation}
		where $H$ stands for harmonic mean, $\mathcal{D}^{\leq t}$ denotes all train/test data from $1^{st}$ to $t^{th}$ task, $\mathcal{D}^{> t}$ denotes all train/test data from $(t+1)^{th}$ to the last task, and $A$ denotes the set of all class attributes. 	 
		
	\end{itemize}
	
	\myparagraph{Limitation of Fixed CZSL Setting:} Since all classes from all tasks are available as a seen or unseen class, the setting cannot be utilized for a class-incremental setup of continual learning. Note that it is an infeasible assumption that all classes' attribute information is known at the first task.
	
	\myparagraph{Dynamic CZSL Setting:}
	
	The new setting is proposed for two reasons: (i) Test the method for class-incremental setup (i.e., dynamical). (ii) Split the data such-wise so that standard seen-unseen classes' split of ZSL dataset \cite{zeroshotlearning_dataset} is maintained at the last task's seen-unseen classes' split of CZSL dataset. In this setting, seen/unseen classes of the standard split are divided among various tasks for the CZSL experiment. If any class is treated as seen/unseen for $t^{th}$ task, those classes will be seen/unseen for all the next tasks. During the evaluation of the model at $t^{th}$ task, testing is performed on current and all previous tasks' test data (but not on any of the next tasks like fixed CZSL setting). Like standard ZSL evaluation, the test data of $t^{th}$ task in CZSL consists of $20\%$ data from seen classes of $t^{th}$ task and unseen classes of $t^{th}$ task. If one follows the above-mentioned setting, the testing data of the last task in CZSL (continual learning) is identical to testing data of standard split in ZSL (offline learning).  Following evaluation metrics are used for the performance analysis of this setting:
	
	\begin{itemize} \itemsep-1.0em
		\item Mean Seen Accuracy (mSA)
		\begin{equation}
		mSA = \frac{1}{T}\sum_{t=1}^T CAcc(\mathcal{D}_{\tau s_s}^{\leq t}, A^{\leq t}_{\tau s_s}),
		\end{equation}
		where $T$ denotes the total number of tasks, and $\mathcal{D}_{\tau s_s}^t$ denotes the test data of $t^{th}$ task from seen classes and $A^{\leq t}_{\tau s_s}$ is the corresponding class attribute of those seen classes.
		\item Mean Unseen Accuracy (mUA)
		\begin{equation}
		mUA = \frac{1}{T}\sum_{t=1}^{T} CAcc(\mathcal{D}_{\tau s_{us}}^{\leq t}, A^{\leq t}_{\tau s_{us}}),
		\end{equation}
		where  $\mathcal{D}_{\tau s_{us}}^t$ denotes the unseen test data of $t^{th}$ task, and $A^{\leq t}_{\tau s_{us}}$ is the corresponding class attribute of those unseen classes.
		\item Mean Harmonic Accuracy (mH)
		\begin{equation}
		mH = \frac{1}{T}\sum_{t=1}^{T} H(\mathcal{D}_{\tau s_s}^{\leq t}, \mathcal{D}_{\tau s_{us}}^{\leq t}, A^{\leq t}),
		\end{equation}
		where $T$ denotes the total number of tasks, $\mathcal{D}_{\tau s_s}^t$ denotes the test data of $t^{th}$ task from seen classes,  $\mathcal{D}_{\tau s_{us}}^t$ denotes the unseen test data of $t^{th}$ task, and $A$ denotes the set of all class attributes. 	
		
	\end{itemize}
	
	It is to be noted that both settings of CZSL are evaluated for generalized zero-shot learning, i.e., generalized CZSL (GCZSL). Description of all datasets' split for fixed and dynamic CZSL setting is provided in the subsequent section.  	
	
	{
		\renewcommand{\arraystretch}{1.4} 
		\begin{table*}[!h]
			\centering
			\resizebox{\textwidth}{!}{%
				\begin{tabular}{|l|ccc|ccc|ccc|ccc|ccc|}
					\hline
					& \multicolumn{3}{c|}{CUB} & \multicolumn{3}{c|}{aPY} & \multicolumn{3}{c|}{AWA1} & \multicolumn{3}{c|}{AWA2} & \multicolumn{3}{c|}{SUN} \bigstrut\\
					\cline{2-16}          & \multicolumn{1}{p{4.215em}}{mSA} & \multicolumn{1}{p{4.215em}}{mUA} & \multicolumn{1}{p{4.215em}|}{mH} & \multicolumn{1}{p{4.215em}}{mSA} & \multicolumn{1}{p{4.215em}}{mUA} & \multicolumn{1}{p{4.215em}|}{mH} & \multicolumn{1}{p{4.215em}}{mSA} & \multicolumn{1}{p{4.215em}}{mUA} & \multicolumn{1}{p{4.215em}|}{mH} & \multicolumn{1}{p{4.215em}}{mSA} & \multicolumn{1}{p{4.215em}}{mUA} & \multicolumn{1}{p{4.215em}|}{mH} & \multicolumn{1}{p{4.215em}}{mSA} & \multicolumn{1}{p{4.215em}}{mUA} & \multicolumn{1}{p{4.215em}|}{mH} \bigstrut\\
					\hline
					SeqL1 & 15.25 & 5.33  & 7.75  & 28.81 & 5.19  & 8.07  & 38.52 & 8.59  & 13.05 & 40.25 & 9.67  & 14.04 & 7.41  & 3.92  & 5.04 \bigstrut[t]\\
					SeqL2 & 24.66 & 8.57  & 12.18 & 46.99 & 13.04 & 19.12 & 44.52 & 15.03 & 21.67 & 47.94 & 16.02 & 23.17 & 16.88 & 11.40 & 13.38 \\
					AGEM+CZSL \cite{skorokhodov2020normalization,chaudhry2018efficient} & --  & --  & 17.30 & --  & --  & --  & --  & --  & --  & --  & --  & --  & --  & --  & 9.60 \\
					EWC+CZSL \cite{skorokhodov2020normalization,schwarz2018progress} & --  & --  & 18.00 & --  & --  & --  & --  & --  & --  & --  & --  & --  & --  & --  & 9.60 \\
					MAS+CZSL \cite{skorokhodov2020normalization,aljundi2018memory} & --  & --  & 17.70 & --  & --  & --  & --  & --  & --  & --  & --  & --  & --  & --  & 9.40 \\														
					\hline
					\textbf{GRCZSL} & \textbf{41.91} & \textbf{14.12} & \textbf{20.48} & \textbf{62.27} & \textbf{12.57} & \textbf{20.46} & \textbf{77.36} & \textbf{23.24} & \textbf{34.86} & \textbf{80.57} & \textbf{24.35} & \textbf{36.57} & \textbf{17.74} & \textbf{11.50} & \textbf{13.73} \bigstrut\\
					\hline
				\end{tabular}%
			}
			\vspace{1mm}
			\caption{Mean seen-class accuracy (mSA), mean unseen-class accuracy (mUA), and their mean of harmonic mean (mH) are given for generalized Fixed CZSL setting. The best results in the table are presented in bold face.}
			\label{tab:gen_res_S1}%
		\end{table*}%
	}	
	
	{
		\renewcommand{\arraystretch}{1.4} 
		\begin{table*}[t]
			\centering
			\resizebox{\textwidth}{!}{%
				\begin{tabular}{|l|ccc|ccc|ccc|ccc|ccc|}
					\hline
					& \multicolumn{3}{c|}{CUB} & \multicolumn{3}{c|}{aPY} & \multicolumn{3}{c|}{AWA1} & \multicolumn{3}{c|}{AWA2} & \multicolumn{3}{c|}{SUN} \bigstrut\\
					\cline{2-16}          & \multicolumn{1}{p{4.215em}}{mSA} & \multicolumn{1}{p{4.215em}}{mUA} & \multicolumn{1}{p{4.215em}|}{mH} & \multicolumn{1}{p{4.215em}}{mSA} & \multicolumn{1}{p{4.215em}}{mUA} & \multicolumn{1}{p{4.215em}|}{mH} & \multicolumn{1}{p{4.215em}}{mSA} & \multicolumn{1}{p{4.215em}}{mUA} & \multicolumn{1}{p{4.215em}|}{mH} & \multicolumn{1}{p{4.215em}}{mSA} & \multicolumn{1}{p{4.215em}}{mUA} & \multicolumn{1}{p{4.215em}|}{mH} & \multicolumn{1}{p{4.215em}}{mSA} & \multicolumn{1}{p{4.215em}}{mUA} & \multicolumn{1}{p{4.215em}|}{mH} \bigstrut\\
					\hline
					SeqL1 & 25.71 & 12.78 & 16.76 & 52.18 & 6.93  & 11.24 & 49.90 & 15.32 & 22.50 & 55.46 & 14.16 & 21.98 & 15.36 & 9.28  & 11.03 \bigstrut[t]\\
					SeqL2 & 38.95 & 20.89 & 26.74 & 61.12 & 14.22 & 22.91 & 63.11 & 30.98 & 40.84 & 67.68 & 35.64 & 46.01 & 29.06 & 21.33 & 24.33 \bigstrut[b]\\
					\hline
					\textbf{GRCZSL} & \textbf{59.27} & \textbf{26.03} & \textbf{35.67} & \textbf{77.14} & \textbf{17.94} & \textbf{28.75} & \textbf{86.92} & \textbf{33.21} & \textbf{47.48} & \textbf{90.61} & \textbf{37.56} & \textbf{52.61} & \textbf{30.78} & \textbf{22.59} & \textbf{25.54} \bigstrut\\
					\hline
				\end{tabular}%
			}	
			\vspace{1mm}
			\caption{The mean seen-class accuracy (mSA), mean unseen-class accuracy (mUA), and their mean of harmonic mean (mH) are calculated for generalized dynamic CZSL setting. The best results in the table are presented in bold face.}
			\label{tab:gen_res_S2}%
		\end{table*}%
	}	
	
	{
		\renewcommand{\arraystretch}{1.0} 
		\begin{table}[!htbp]
			\centering
			\begin{tabular}{|l|c|c|c|c|c|}
				\hline
				& CUB   & aPY   & AWA1  & AWA2  & SUN \bigstrut\\
				\hline
				\multicolumn{1}{|p{5.715em}|}{Offline (Upper Bound)} & 34.50 & 22.38 & 47.20 & 51.20 & 26.7 \bigstrut\\
				\hline
				SeqL1 & 12.17 & 0.72  & 20.70 & 16.26 & 8.21 \bigstrut[t]\\
				SeqL2 & 14.89 & 14.11  & 16.29 & 25.36 & 16.85 \bigstrut[b]\\
				\hline
				\textbf{GRCZSL} & 26.34 & 17.68 & 37.06 & 38.71 & 20.80 \bigstrut\\
				\hline
			\end{tabular}%
			\vspace{1mm}
			\caption{Performance comparison of the CZSL methods at the last task and their corresponding offline model, i.e., when we present all tasks training and testing data at once}
			\label{tab:batch_vs_online}%
		\end{table}%
	}
	
	\subsection{Dataset Division as Per Fixed and Dynamic Settings for CZSL}
	
	As mentioned in the above section, two kinds of settings are used for the experiments. Datasets' split as per both settings are mentioned below:
	
	\myparagraph{For fixed CZSL setting:} 
	The $200$ classes of the CUB dataset are split into $20$ tasks of $10$ classes each. Similarly, the aPY dataset which contains $32$ classes is split into $8$ tasks with $4$ classes each. For the AWA1 and AWA2 datasets which have $50$ classes each are split into $10$ tasks with $5$ classes each. The SUN dataset has $717$ classes and is difficult to split evenly. Hence, it is split into $15$ tasks with $47$ classes in the first $3$ tasks and $48$ classes in the remaining tasks. For all datasets, $20$ percent of data from each task is taken as test data to compute the final evaluation metrics.
	
	\myparagraph{For dynamic CZSL setting:}
	The CUB dataset is split into 20 tasks with first 10 tasks containing 7 seen classes and 3 unseen classes each and next 10 tasks containing 8 seen classes and 2 unseen classes each. Here, the test data consists of the unseen classes and 20 percent data from seen classes of each task. The aPY dataset splits into 8 tasks with first 4 tasks containing 2 seen classes and 2 unseen classes each and remaining tasks containing 3 seen classes and 1 unseen classes each. The AWA1 and AWA2 split into 10 tasks with 4 seen classes and 1 unseen classes per task. The SUN dataset splits into 15 tasks with first 3 tasks containing 43 seen classes and 4 unseen classes each and remaining tasks containing 43 seen classes and 5 unseen classes each.

	\subsection{Comparative Analysis}
	
	There are only a handful of research works available on continual zero-shot learning. We found three works so far, which can be categorized under multi-head \cite{chaudhry2018efficient,wei2020lifelong} and single-head settings \cite{skorokhodov2020normalization}. Since the single-head setting is more feasible for the practical scenario when compared to the multi-head setting	\cite{chaudhry2018riemannian}, the single-head setting is used for experiment evaluation more practical and challenging. Hence, we have used only the single-head setting-based method \cite{skorokhodov2020normalization} for a fair comparison. Following baselines are considered for comparative study.                                                                                                                         
	\begin{itemize}
		\item \textbf{AGEM+CZSL \cite{skorokhodov2020normalization}:} 
		In \cite{skorokhodov2020normalization}, authors developed various CZSL methods based on average gradient episodic memory \cite{chaudhry2018efficient}, elastic weight consolidation \cite{kirkpatrick2017overcoming,schwarz2018progress}, and memory aware synapses \cite{aljundi2018memory} for single-head setting using fixed CZSL setting. Since, the authors presented results on only two datasets, namely CUB and SUN datasets, results on these two datasets are provided in Table \ref{tab:gen_res_S1} for comparison.     
		
		\item \textbf{Sequential baselines:} For developing baselines for CZSL, we sequentially train the base classifier with CVAE over multiple tasks without considering any continual learning strategy. In sequential training, initial weights of the CVAE at $t^{th}$ task is the final weight of CVAE at $(t-1)^{th}$ task. After training the Seq-CVAE on the current task, synthetic samples are generated using trained decoder ($D_{t}$) for all the classes which need to classify by using class attribute information. Here, two types of regularizations (L1 and L2 regularizations) have been utilized to generate two baseline methods (SeqL1 and SeqL2).  
	\end{itemize}
	
	The extensive experiments have been performed based on both the settings for GRCZSL, and performance analysis is given below:  
	
	\myparagraph{GRCZSL for Fixed Setting.}
	Results for this setting are provided in Table \ref{tab:gen_res_S1} for all $5$ datasets. For the CUB dataset, GRCZSL significantly outperforms the two baselines by more than $8\%$ in terms of mH. It yields more than $3\%$, $2\%$, and $2\%$ mH value compared to the state of the art CZSL methods AGEM+CZSL \cite{skorokhodov2020normalization}, EWC+CZSL \cite{skorokhodov2020normalization,kirkpatrick2017overcoming}, and MAS+CZSL \cite{skorokhodov2020normalization,aljundi2018memory}, respectively. Similarly, GRCZSL improves the performance over the baselines by more than $11\%$ and $12\%$ for AWA1 and AWA2 datasets, respectively. These improvements have been gained due to replay the synthetic samples generated by the immediately previous model. Here, we do not need to store any samples from the real data explicitly. At any point of time during training for $t^{th}$ task, GRCZSL requires only two models, i.e., one is for the current task, and the other is for the previous task. The proposed method exhibits the least mH improvements over baselines on the aPY and SUN datasets by more than $1.5\%$ and $0.3\%$, respectively.    
	
	\myparagraph{GRCZSL for Dynamic Setting.} 
	Results for this setting are provided in Table \ref{tab:gen_res_S2} for all $5$ datasets. GRCZSL outperforms all baseline methods for all datasets. It exhibits a significant improvement of more than $8\%$, $6\%$, $6\%$, $6\%$, and $1\%$ for CUB, aPY, AWA1, AWA2, and SUN datasets, respectively.  As mentioned above, for this setting, the performance on the last task should be equal to the base offline method CVAE, which is the upper bound for our proposed method. The mH values on the last task for all methods are presented in Table \ref{tab:batch_vs_online}. It can be analyzed from this table that GRCZSL exhibits better performance than baselines; however, there is still room for improvement as GRCZSL lacks behind the upper bound.

	%
	
	\begin{figure*}[h!]
		\centering
		\begin{subfigure}{.5\textwidth}
			\centering
			\includegraphics[width=1.1\linewidth]{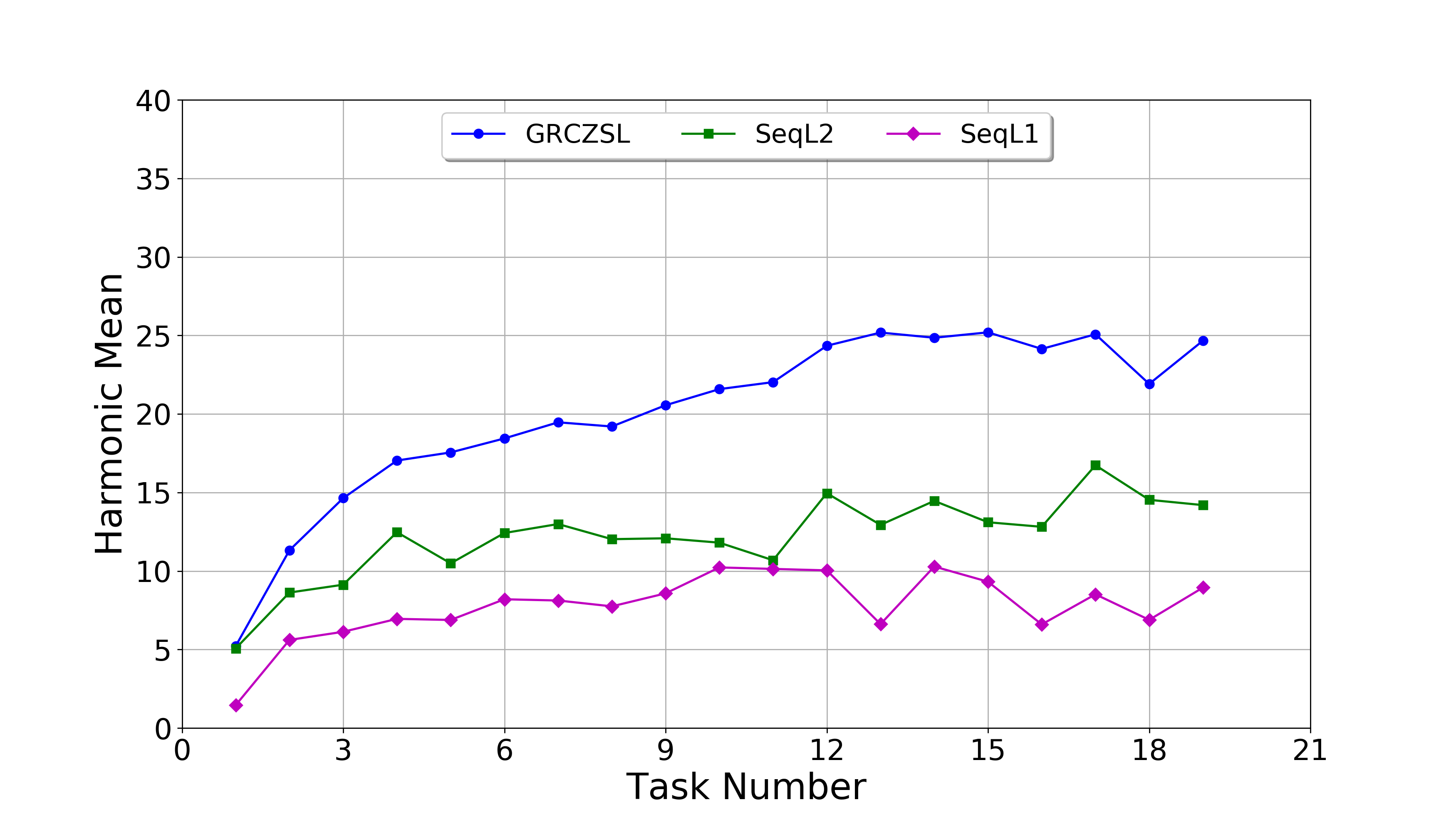}
			\caption{For fixed CZSL setting}
			\label{fig:cub-mH_s1}
		\end{subfigure}
		\begin{subfigure}{.49\textwidth}
			\centering
			\includegraphics[width=1.1\linewidth]{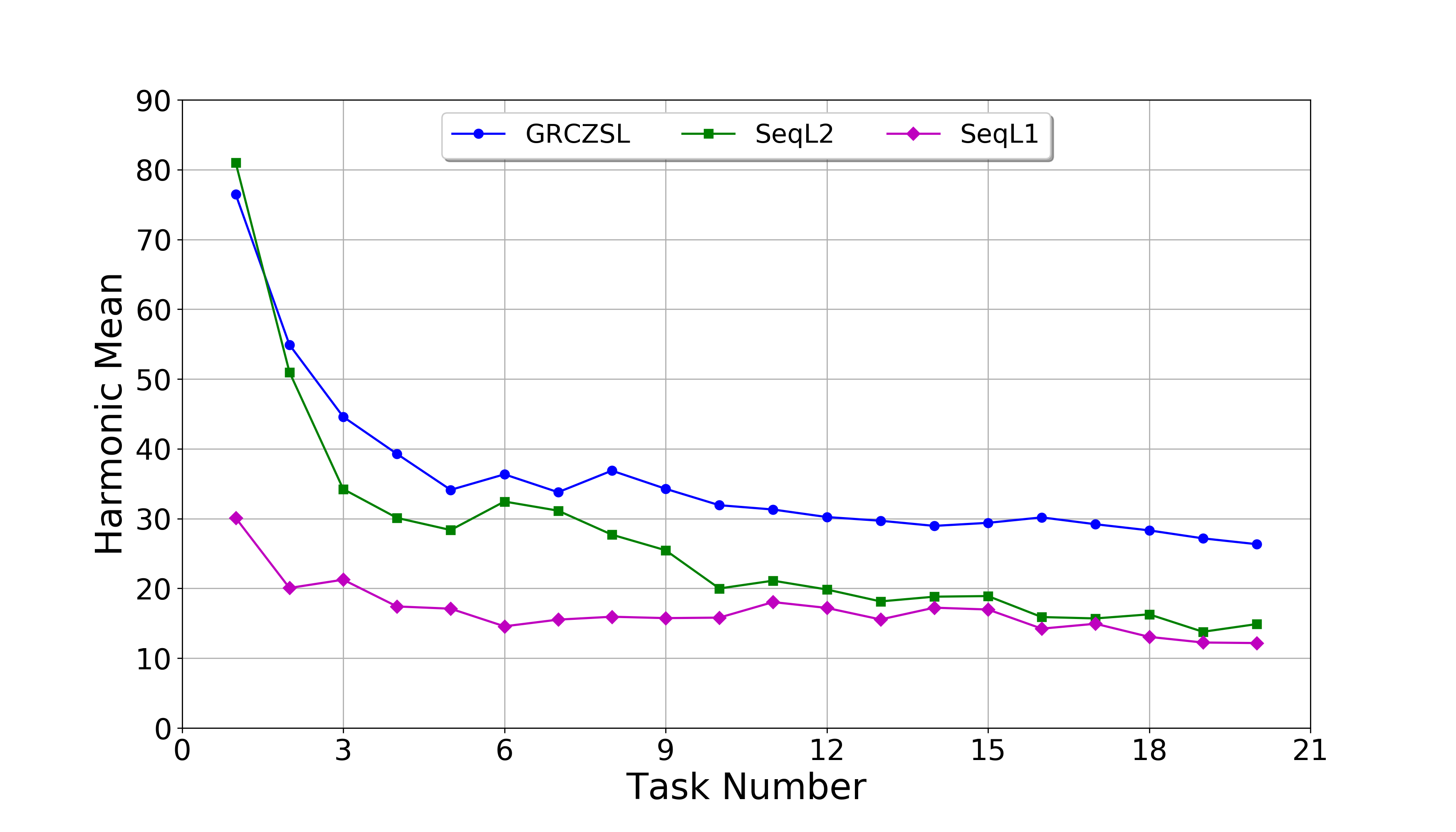}
			\caption{For dynamic CZSL setting}
			\label{fig:cub-mH_s2}
		\end{subfigure}
		\caption{Harmonic mean for CUB dataset over $20$ tasks for fixed and dynamic CZSL settings.}
		\label{fig:taskwise-mH}
	\end{figure*}

	
	\begin{figure*}
		\centering
		\begin{subfigure}{.5\textwidth}
			\centering
			\includegraphics[width=1.1\linewidth]{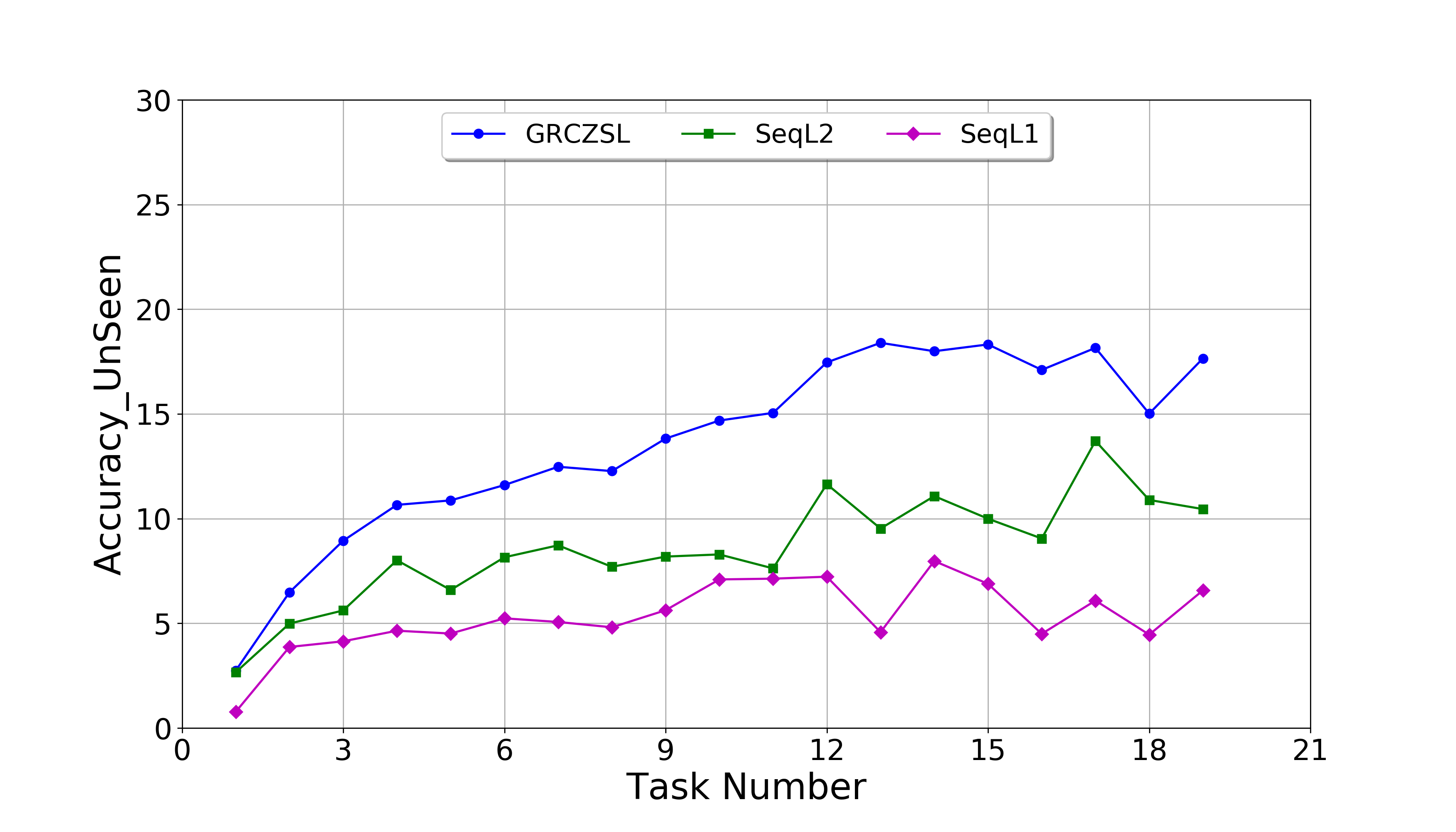}
			\caption{For fixed CZSL setting}
			\label{fig:cub-mUA_s1}
		\end{subfigure}
		\begin{subfigure}{.49\textwidth}
			\centering
			\includegraphics[width=1.1\linewidth]{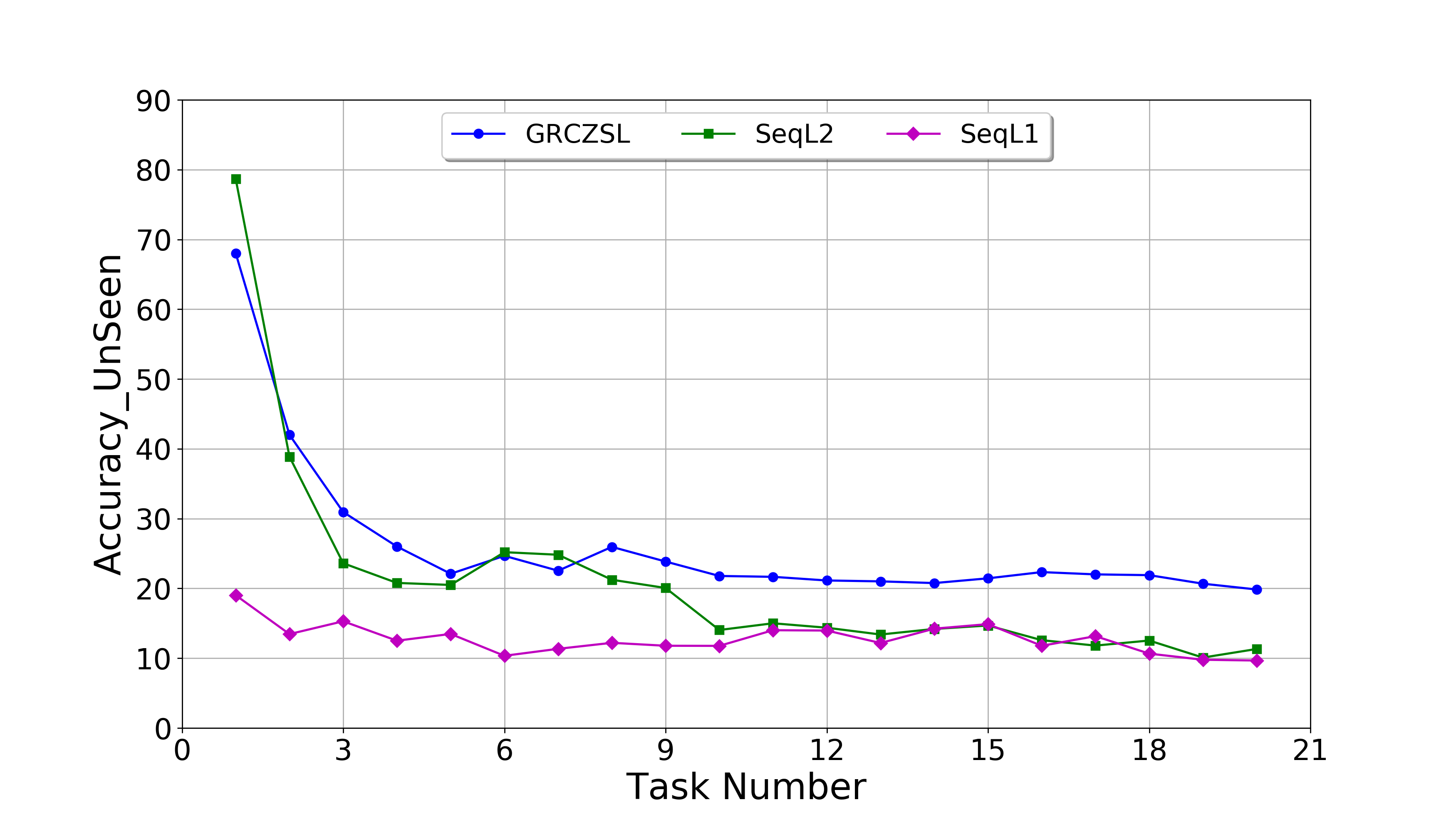}
			\caption{For dynamic CZSL setting}
			\label{fig:cub-mUA_s2}
		\end{subfigure}
		\caption{Per class unseen accuracy for CUB dataset over $20$ tasks for fixed and dynamic CZSL settings.}
		\label{fig:taskwise-mUA}
	\end{figure*}

	

	
	\begin{figure*}[h!]
		\centering
		\begin{subfigure}{.5\textwidth}
			\centering
			\includegraphics[width=1.1\linewidth]{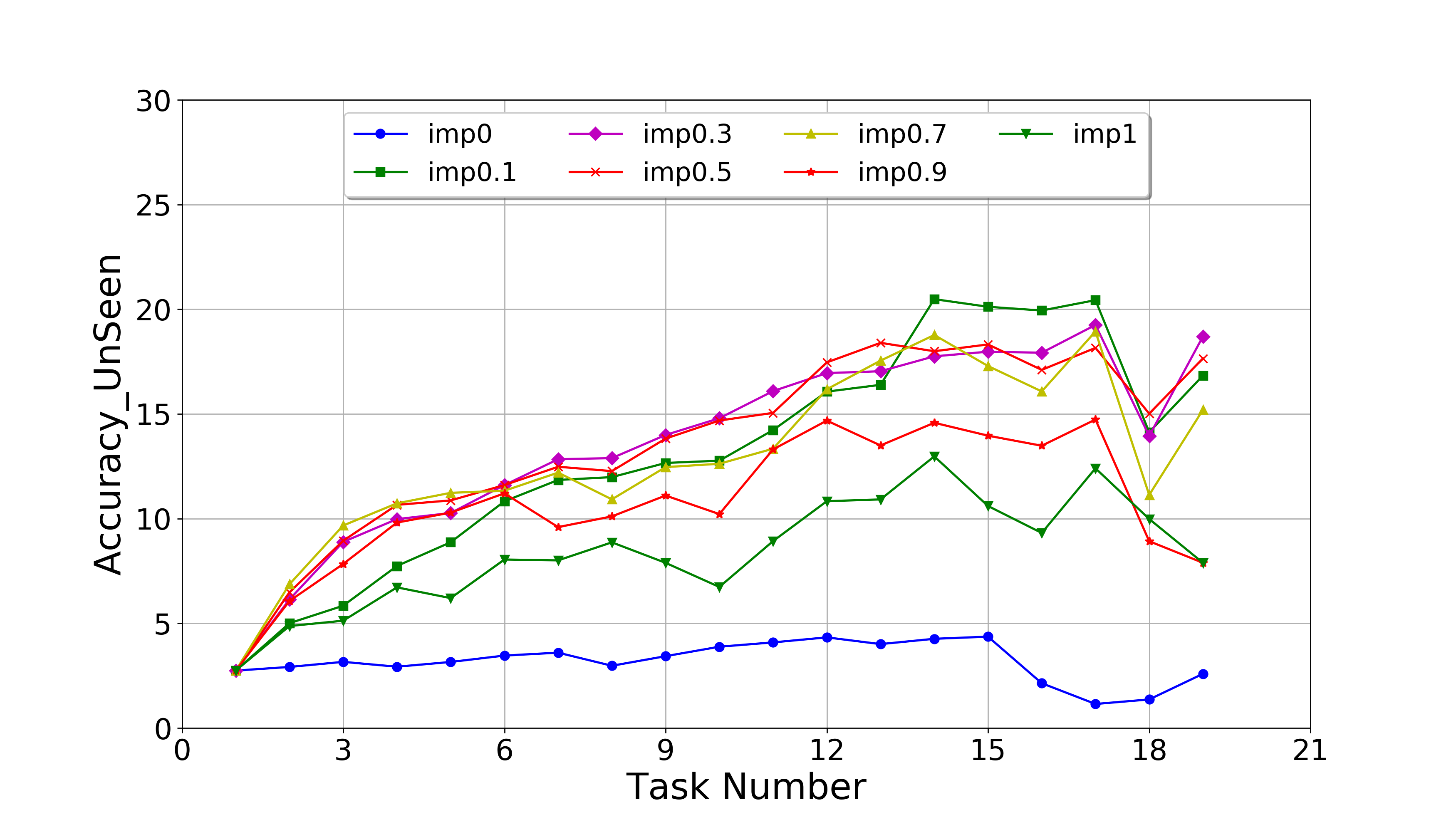}
			\caption{For fixed CZSL setting}
			\label{fig:cub-mUA_s1_Imp}
		\end{subfigure}
		\begin{subfigure}{.49\textwidth}
			\centering
			\includegraphics[width=1.1\linewidth]{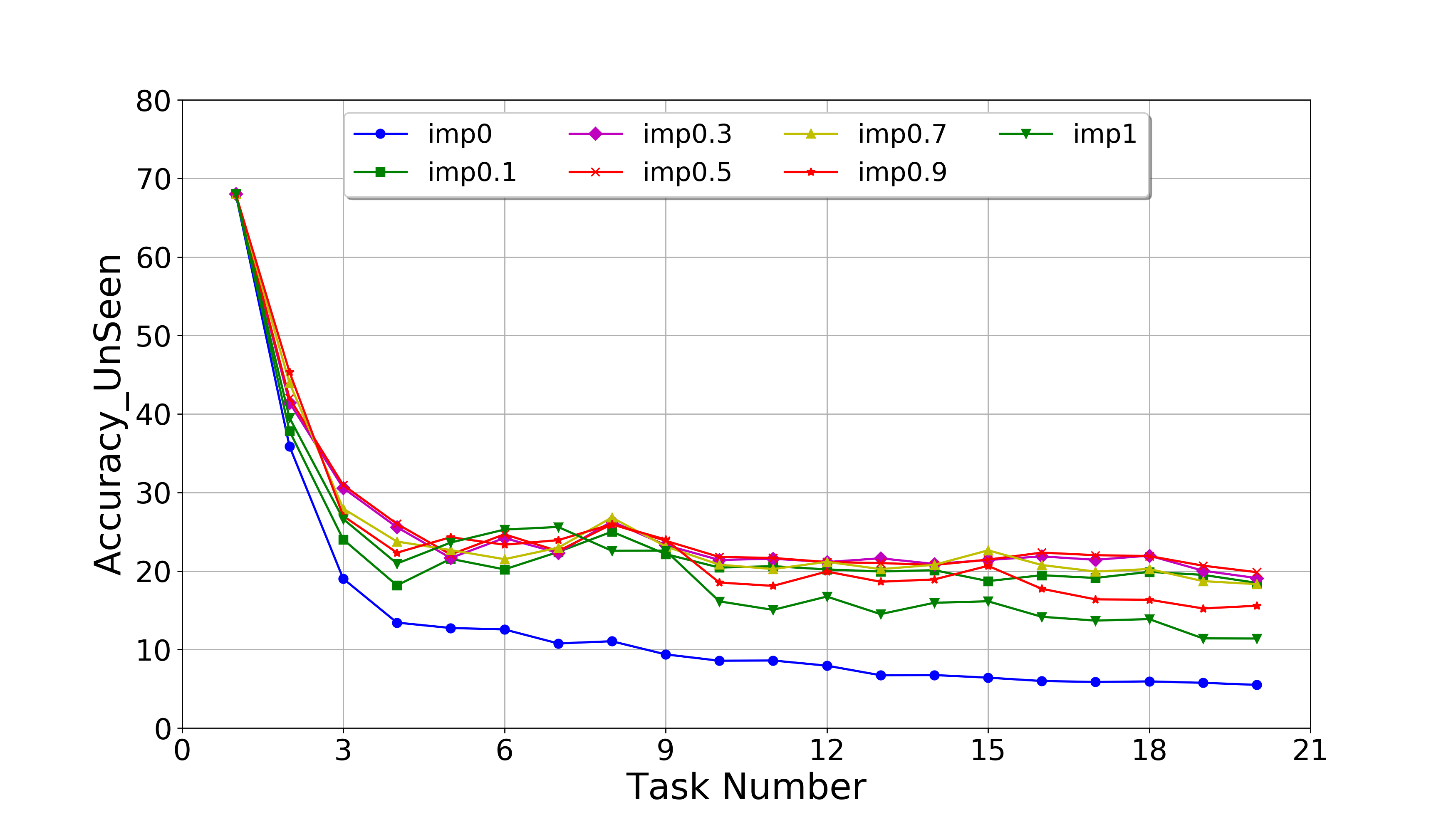}
			\caption{For dynamic CZSL setting}
			\label{fig:cub-mUA_s2_Imp}
		\end{subfigure}
		\caption{Per class unseen accuracy as per task importance for CUB dataset over $20$ tasks for fixed and dynamic CZSL settings.}
		\label{fig:taskwise-mUA_Imp}
	\end{figure*}

	\begin{figure*}[h!]
		\centering
		\begin{subfigure}{.5\textwidth}
			\centering
			\includegraphics[width=1.1\linewidth,height=5cm]{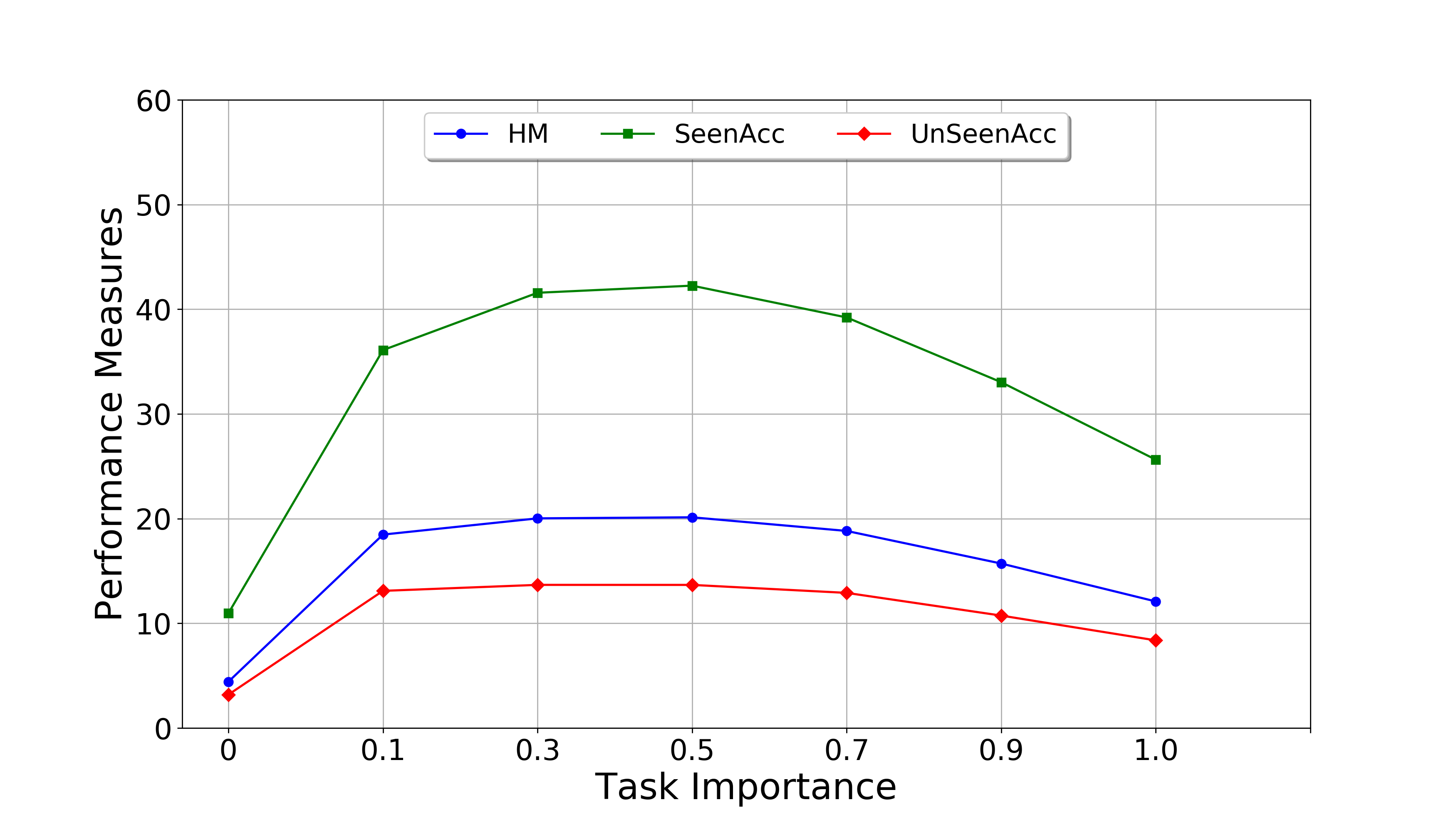}
			\caption{For fixed CZSL setting}
			\label{fig:abal-taskimp_s1}
		\end{subfigure}
		\begin{subfigure}{.49\textwidth}
			\centering
			\includegraphics[width=1.1\linewidth,height=5cm]{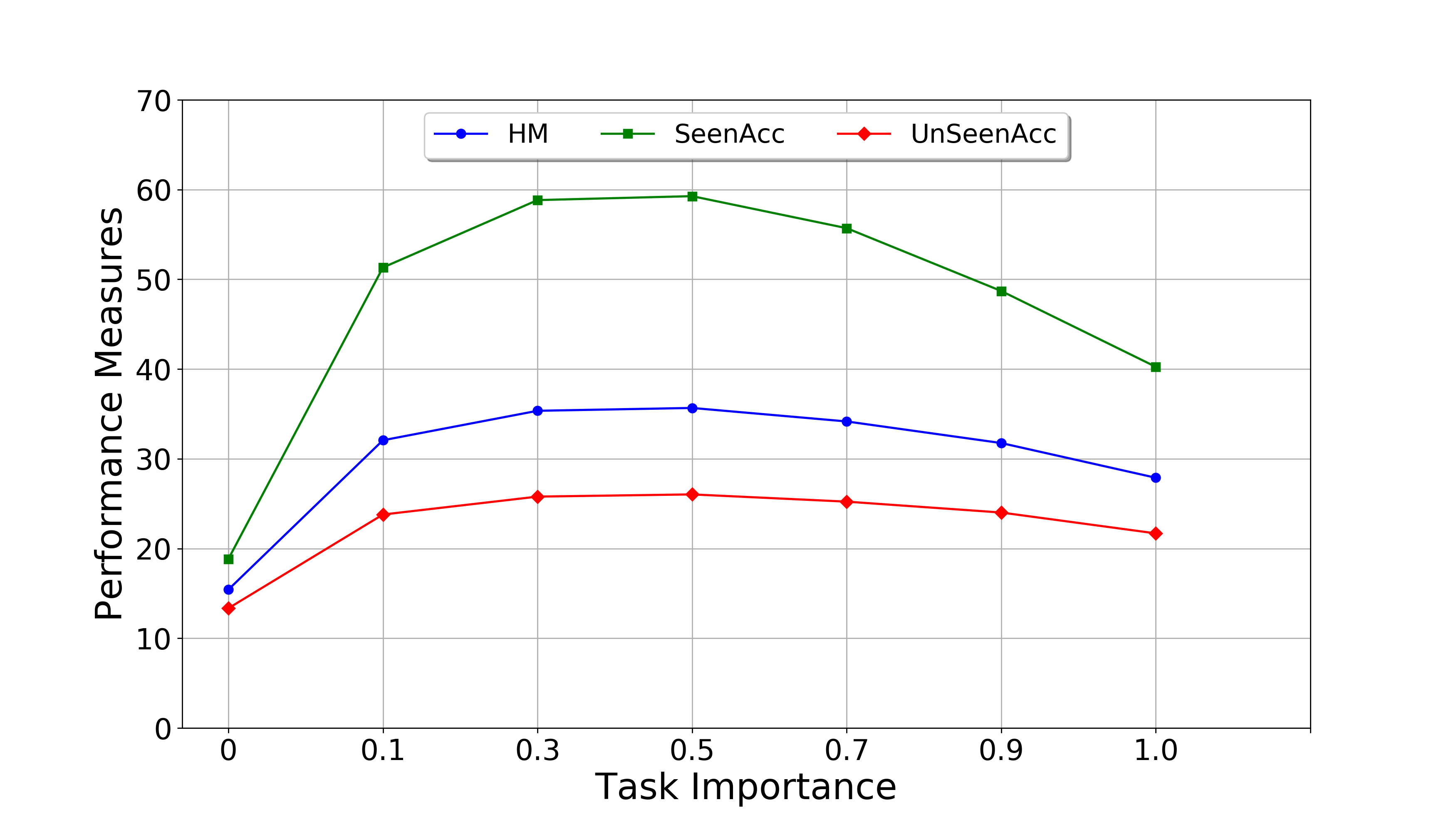}
			\caption{For dynamic CZSL setting}
			\label{fig:abal-taskimp_s2}
		\end{subfigure}
		\caption{The impact of the task importance on CUB dataset in terms of mH, mSA, and mUA for both settings.}
		\label{fig:abal-taskimp}
	\end{figure*}
	
	\clearpage
	\subsection{Ablation Study on CUB dataset}
	
	The ablation study is conducted using CUB dataset.
	
	\myparagraph{Task-wise Performance Analysis:}
	Task-wise results are plotted in Figure \ref{fig:taskwise-mH} in term of mH. These plots are generated by providing equal importance to the samples from the current and previous tasks. We observed that overall performance (i.e., mH) and performance on unseen classes (i.e., mUA) follow a similar pattern over $20$ tasks of the CUB dataset for each setting (please refer to Table \ref{fig:taskwise-mUA} for the plot of mUA).  We observe this pattern due to the lower performance of unseen classes, which is the main reason for the lower performance of any model in zero-shot learning. However, fixed and dynamic settings exhibit different patterns in \ref{fig:taskwise-mH} because the number of unseen classes is decreasing, and the number of seen classes increases with each task in the case of fixed CZSL setting. In contrast, the number of seen and unseen classes is constantly increasing with each task in the case of dynamic CZSL setting. Therefore, we observed these kinds of different patterns in both settings.    
	
	\myparagraph{Task Importance Analysis:}  
	For analyzing the importance of the current and previous tasks, task importance ($\alpha$) is varied in the range $\alpha= [0, 0.1, 0.3, 0.5, 0.7, 0.9, 1]$. Like the above discussion, the same pattern is observed between overall performance and the performance on unseen classes over $20$ tasks of the CUB dataset (see Figure \ref{fig:taskwise-mUA_Imp} for the plots). It can be analyzed from Figure \ref{fig:abal-taskimp} that GRCZSL performed worst when $\alpha=0$, i.e., there is no importance of the current task. The second worst performance is observed when $\alpha=1$, i.e., there is no importance of previous tasks. The best performance is observed when task importance is in the range $[0.3, 0.5]$ for both settings.      
	
	
	\section{Conclusion} \label{concl}
	The presented a deep generative replay-based continual zero-shot learning (GRCZSL) method for a single-head setting, capable of handling unseen classes by only using semantic information. The GRCZSL only requires an entire network for the current task and a previously trained decoder alone from the immediate previous task. The network performs CZSL, and the previous (only decoder) trained network alleviates the catastrophic forgetting by generating replay samples for the classes of the previously learned tasks. These samples help the network in retaining the knowledge of the earlier tasks. GRCZSL also provides a balance between the former and the new task's performance. Overall, the proposed GRCZSL can be used in the continual/lifelong learning setting without retraining from scratch. Experimental results on five datasets and ablation study on the CUB dataset also exhibited that GRCZSL significantly outperformed the baseline and the existing state-of-the-art methods. The current CZSL evaluation setting problem is alleviated in the paper by new CZSL class incremental and task-free learning settings. The performance of GRCZSL has been evaluated in the setting, and it is compared with the baseline. Since the GRCZSL requires task information during training, it is not ideal for task-free learning. Therefore, it is necessary to develop a CZSL method, which will be appropriate for both class-incremental and task-free learning.

	
	{\small
		\bibliographystyle{unsrt}
		\bibliography{egbib}
	}
\end{document}